\documentclass[lettersize,journal]{IEEEtran}
\usepackage{amsmath,amsfonts}
\usepackage{algorithm}
\usepackage{algorithmic}
\usepackage{array}
\usepackage[caption=false,font=normalsize,labelfont=sf,textfont=sf]{subfig}
\usepackage{textcomp}
\usepackage{stfloats}
\usepackage{url}
\usepackage{verbatim}
\usepackage{graphicx}
\usepackage{cite}
\usepackage{multirow}
\usepackage{booktabs}
\usepackage{amssymb}
\usepackage{amsmath,amsfonts}
\usepackage[hidelinks]{hyperref}
\usepackage{newtxtext}
\usepackage{newtxmath}
\def\BibTeX{{\rm B\kern-.05em{\sc i\kern-.025em b}\kern-.08em
    T\kern-.1667em\lower.7ex\hbox{E}\kern-.125emX}}
\usepackage{balance}
\begin{document}

\title{Dual-Layer Semantic-Spatial Belief Mapping\\
for Aerial Object Goal Navigation}

\author{Jianqiang Xiao, Xiang Deng, Yuexuan Sun, Yanjin Wu, Wenbiao Yan,~and Liqiang Nie*%
\thanks{Manuscript received XXXX; revised XXXX.}%
}

\markboth{IEEE Transactions on Multimedia}%
{Author1 et al.: Dual-Layer Semantic-Spatial Belief Mapping for Aerial Object Goal Navigation}

\maketitle

\begin{abstract}
Aerial Object Goal Navigation (ObjectNav) requires an unmanned aerial vehicle (UAV) to locate a described target in an unknown outdoor environment using onboard visual observations. Vision-language models (VLMs) provide a natural means of interpreting open-ended target descriptions and visual observations, but their frame-level outputs are often noisy, sparse, and spatially transient, limiting their effectiveness for sustained autonomous search. We propose AeroBelief, a dual-layer semantic-spatial belief mapping framework that transforms transient VLM observations into persistent spatial guidance. The framework separates broad contextual plausibility from target-specific evidence. An intuition layer accumulates scene-level semantic cues for exploration, while an evidence layer preserves qualified target-specific observations for approach and confirmation. Evidence-gated fusion combines the two layers into spatial belief hotspots for planner-level decision making. To improve observation reliability before spatial accumulation, we further introduce object-conditioned visual reasoning with conservative evidence qualification, which assesses candidate target-specific observations according to contextual consistency and target-level verification requirements. In parallel, egocentric regional guidance converts quadtree coverage into UAV-centered, yaw-aligned directional proposals and stabilizes them through temporal commitment across steps. Its regional scoring operates independently of semantic belief values, thereby maintaining exploration pressure, reducing repeated low-gain search, and limiting the premature dominance of local semantic hotspots. Experiments on the UAV-ON benchmark show that AeroBelief achieves the best reported overall SR, OSR, and SPL among the compared methods, reaching 21.61\%, 35.57\%, and 10.62, respectively. These results support the effectiveness of persistent semantic-spatial belief, conservative evidence qualification, and temporally stable regional guidance for aerial ObjectNav. Code and implementation details are publicly available at \url{https://github.com/Shawnjx/AeroBelief}.
\end{abstract}


\section{Introduction}
\IEEEPARstart{A}{utonomous} unmanned aerial vehicles (UAVs) are increasingly expected to operate in large-scale, open, and partially unknown environments for applications such as infrastructure inspection~\cite{uav_inspection}, patrol and monitoring~\cite{uav_patrol}, search and rescue~\cite{uav_sar}, and aerial target tracking~\cite{uav_target_tracking}. A fundamental capability in these scenarios is aerial object goal navigation (ObjectNav)~\cite{uavon}, where a UAV searches for and approaches a user-specified target based on a high-level semantic description rather than following a predefined waypoint sequence. Compared with ground-based ObjectNav~\cite{habitat} and instruction-following vision-language navigation~\cite{r2r,AerialVLN}, this setting presents several distinctive challenges. The target location is unknown, the observation viewpoint varies with both horizontal motion and altitude, and the UAV needs to reason over large outdoor spaces characterized by occlusion, sparse target visibility, and visually similar background structures.

Existing embodied navigation research has advanced along complementary directions in spatial exploration and semantic reasoning. Map-based exploration, modular planning, and hierarchical policy learning support systematic exploration and goal-directed navigation~\cite{yamauchi,gose,skill_nav}. ObjectNav methods further incorporate scene priors, learned potential functions, and hierarchical object--zone relations to estimate promising search regions when the target is not directly observable~\cite{scene_priors,poni,hozpp}. In parallel, vision-language navigation and recent LLM/VLM-based agents employ language-conditioned reasoning to align visual observations with instructions, maintain navigation context, and generate navigation decisions~\cite{latent_vln,navgpt,mapgpt,navcot}. Collectively, these studies suggest that effective navigation benefits from coupling semantic interpretation with persistent spatial memory. Adapting these mechanisms to aerial ObjectNav, however, remains nontrivial because the agent receives an open-ended target description rather than route-level guidance and needs to infer both promising search regions and informative aerial viewpoints in a large outdoor environment with limited structural constraints.

Three challenges are particularly relevant. First, VLM outputs are typically frame-level semantic judgments with limited spatial persistence. A positive visual cue may disappear from subsequent observations as the UAV moves or changes altitude, and its contribution to later decisions can be lost unless it is preserved within a persistent spatial representation. Second, VLMs may blur the distinction between contextual plausibility and target-specific evidence. A building facade, for example, may provide a plausible context for an air conditioner, yet its presence alone does not constitute evidence that the target is visible. Treating contextual cues and target evidence in the same manner can generate misleading spatial hotspots and repeatedly draw the planner toward irrelevant locations. Third, exploration decisions made independently at each step may exhibit poor temporal consistency in large aerial environments. In the absence of stable coverage-oriented guidance, the UAV may repeatedly revisit explored areas, oscillate among similar directions, or concentrate excessively on local semantic cues while leaving substantial portions of the environment underexplored.

To address these challenges, we propose AeroBelief, a semantic-spatial reasoning framework that transforms transient VLM judgments into persistent target-related spatial belief and stable exploration guidance. Object-conditioned visual reasoning derives target-specific cues and applies conservative evidence qualification before spatial accumulation. A dual-layer belief map maintains contextual intuition and qualified target evidence in separate layers, which are integrated through evidence-gated fusion to form spatial belief hotspots. In parallel, egocentric regional guidance projects quadtree coverage into UAV-centered directional bins and stabilizes the resulting proposal through temporal commitment. The two streams provide complementary information, with semantic belief supporting target-oriented reasoning and regional guidance maintaining systematic exploration when reliable evidence remains sparse. The planner jointly considers fine-grained belief hotspots and coarse coverage-oriented guidance, while a shared UAV execution backbone handles motion feasibility and safety.

The main contributions of AeroBelief are summarized as follows.

\begin{itemize}
\item We propose a dual-layer semantic-spatial belief mapping mechanism for aerial ObjectNav. The intuition layer accumulates broad contextual cues with temporal forgetting, while the evidence layer preserves qualified target-specific observations. Evidence-gated fusion integrates the two layers to form spatial belief hotspots for planner-level decision making.
\item We introduce an object-conditioned visual reasoning module with conservative evidence qualification. Target-conditioned cues capture discriminative attributes, contextual priors, and target-level verification requirements. Scene-context, core-evidence, and part--whole qualification restrict insufficiently supported target claims before they enter persistent spatial memory.
\item We present exploration-aware egocentric regional guidance with temporal commitment. The module projects quadtree coverage into UAV-centered, yaw-aligned directional bins whose scores are computed independently of semantic belief values. Temporal commitment stabilizes the resulting exploration proposal across steps, helping reduce directional oscillation and redundant revisitation without relying on fixed world-space partitions or directly prescribing UAV actions.
\end{itemize}

\section{Related Work}

\subsection{Aerial Embodied Navigation}

Aerial vision-language navigation extends conventional VLN to UAV agents operating in large-scale outdoor environments, where altitude variation, long-range observation, and three-dimensional motion introduce additional challenges. AerialVLN introduced an early city-scale benchmark~\cite{AerialVLN}. TravelUAV developed a realistic simulation platform for instruction-guided UAV navigation~\cite{traveluav}, while OpenFly introduced automated data generation and large-scale aerial VLN evaluation~\cite{openfly}. BEDI broadened the scope to embodied UAV perception, planning, and action generation~\cite{bedi}. CityNav and GeoText-1652 further explored landmark- and geography-aware aerial navigation~\cite{citynav,geotext}. Recent LLM/VLM-based agents have investigated semantic planning, trajectory memory, geospatial reasoning, history-aware control, and open-vocabulary grounding~\cite{flightgpt,geonav,skyvln,grounded_uav_vln}. These studies predominantly follow an instruction-guided paradigm in which language provides route, landmark, geographic, or destination cues.

UAV-ON~\cite{uavon} instead formulates open-world aerial ObjectNav, where the UAV receives an instance-level semantic target description and autonomously determines where and how to search. The target may remain outside the current field of view for extended portions of an episode, making persistent spatial memory and systematic exploration particularly important. APEX combines dynamic spatio-semantic mapping, reinforcement-learning-based action selection, and open-vocabulary target grounding~\cite{apex}, while OctMem-Agent organizes RGB-D observations within an adaptive octree memory to support task-relevant retrieval and frontier-aware exploration~\cite{octmem}. Although these methods demonstrate the value of structured spatial memory, they do not explicitly separate broad contextual plausibility from qualified target-specific evidence before spatial accumulation. We address this distinction through object-conditioned visual reasoning with conservative evidence qualification and dual-layer semantic-spatial belief mapping, while egocentric regional guidance provides stable coverage-oriented exploration independently of semantic belief values.

\subsection{Object Goal Navigation}

Object goal navigation (ObjectNav) considers an embodied agent that searches for an instance of a specified object category in an unseen environment without route-level instructions. Habitat provides a widely used photorealistic simulation platform and evaluation infrastructure for embodied navigation research~\cite{habitat}. Learning-based approaches learn navigation policies from egocentric observations and goal labels through reinforcement learning, meta-learning, or hierarchical visual priors~\cite{target_driven_nav,savn,progressive_nav}. Modular approaches combine semantic mapping, goal prediction, and local planning~\cite{gose}, while generative mapping predicts semantic content in unobserved regions to support target search~\cite{imagine_before_go}. When the target is not directly observable, existing methods infer promising search regions using object--scene priors~\cite{scene_priors}, learned spatial potential functions~\cite{poni}, or hierarchical object--zone relations~\cite{hozpp}. Subsequent studies distinguish target search from target-directed navigation~\cite{dat}, establish competitive modular baselines~\cite{stubborn}, and examine the deployment of ObjectNav systems in real-world environments~\cite{real_world_objectnav}.

Most ObjectNav methods focus on indoor ground agents and assume predefined object categories supported by structured perception models or learned priors. Aerial ObjectNav instead involves three-dimensional motion, pose- and altitude-dependent outdoor observations, and open-ended target descriptions whose VLM interpretations may mix contextual plausibility with target-specific evidence. We explicitly separate these signals through distinct intuition and evidence layers rather than representing them within a single target-likelihood map.

\subsection{Semantic Mapping for Active Search}

Semantic mapping and spatial memory provide persistent representations for navigation in partially observed environments. Occupancy-grid mapping represents free, occupied, and unknown space for geometric planning~\cite{occupancy_grid}, while frontier-based exploration identifies boundaries between known and unknown regions to support systematic coverage~\cite{yamauchi}. Dense semantic mapping associates visual predictions with reconstructed geometry to form semantically enriched scene representations~\cite{semanticfusion}. In aerial settings, active metric--semantic mapping jointly considers geometric and semantic uncertainty to guide multi-robot exploration~\cite{active_metric_semantic_mapping_uav}. Embodied navigation methods further employ learned metric mapping and global planning~\cite{active_neural_slam}, topological spatial memory~\cite{neural_topological_slam}, and semantic prediction beyond observed regions~\cite{learning_to_map}. For active object search, probabilistic and relational memories encode target likelihoods, semantic dependencies, and landmark associations to guide subsequent exploration~\cite{brm,slim}.

These approaches generally rely on structured semantic inputs such as detector outputs, segmentation probabilities, or fixed-category priors. VLM judgments are sparser and more semantically uncertain, and contextual compatibility does not necessarily indicate target presence. Aggregating such signals within a single representation may therefore produce unreliable spatial hotspots. We address this issue by separating contextual intuition from target-specific evidence through dual-layer belief mapping and evidence-gated fusion. In parallel, egocentric regional guidance uses quadtree coverage and temporal commitment to provide stable coverage-oriented guidance without introducing fixed world-space partitions or a separate global planner.

\section{Proposed Method}

\subsection{Task Formulation}

We consider aerial ObjectNav in an unknown outdoor environment. At the beginning of each episode, the UAV receives a target specification
\begin{equation}
g = (n_g, s_g, d_g),
\end{equation}
where $n_g$ denotes the target name, $s_g$ denotes its size category, and $d_g$ denotes a natural-language description. The target position is not provided, and the UAV explores the environment using onboard visual observations.

Let $T$ denote the terminal timestep. Success is recorded when the UAV explicitly issues a \texttt{stop} action within a predefined success radius,
\begin{equation}
S =
\mathbb{1}
\left[
a_T = \texttt{stop}
\ \wedge\
\operatorname{dist}(\mathbf{x}_T,\mathbf{x}_g) \le \delta
\right],
\end{equation}
where $\mathbf{x}_T$ denotes the UAV position at termination, $\mathbf{x}_g$ denotes the target position, and $\delta$ is the success radius. This criterion distinguishes successful stopping from merely entering the target vicinity.

For controlled comparisons among the variants of our framework, all variants share the same UAV execution backbone, including the action space, collision avoidance, boundary handling, altitude regulation, fallback recovery, and stop verification. Further implementation details are provided in the supplementary material.

\subsection{System Overview}

\begin{figure*}[!t]
\centering
\includegraphics[width=0.98\textwidth]{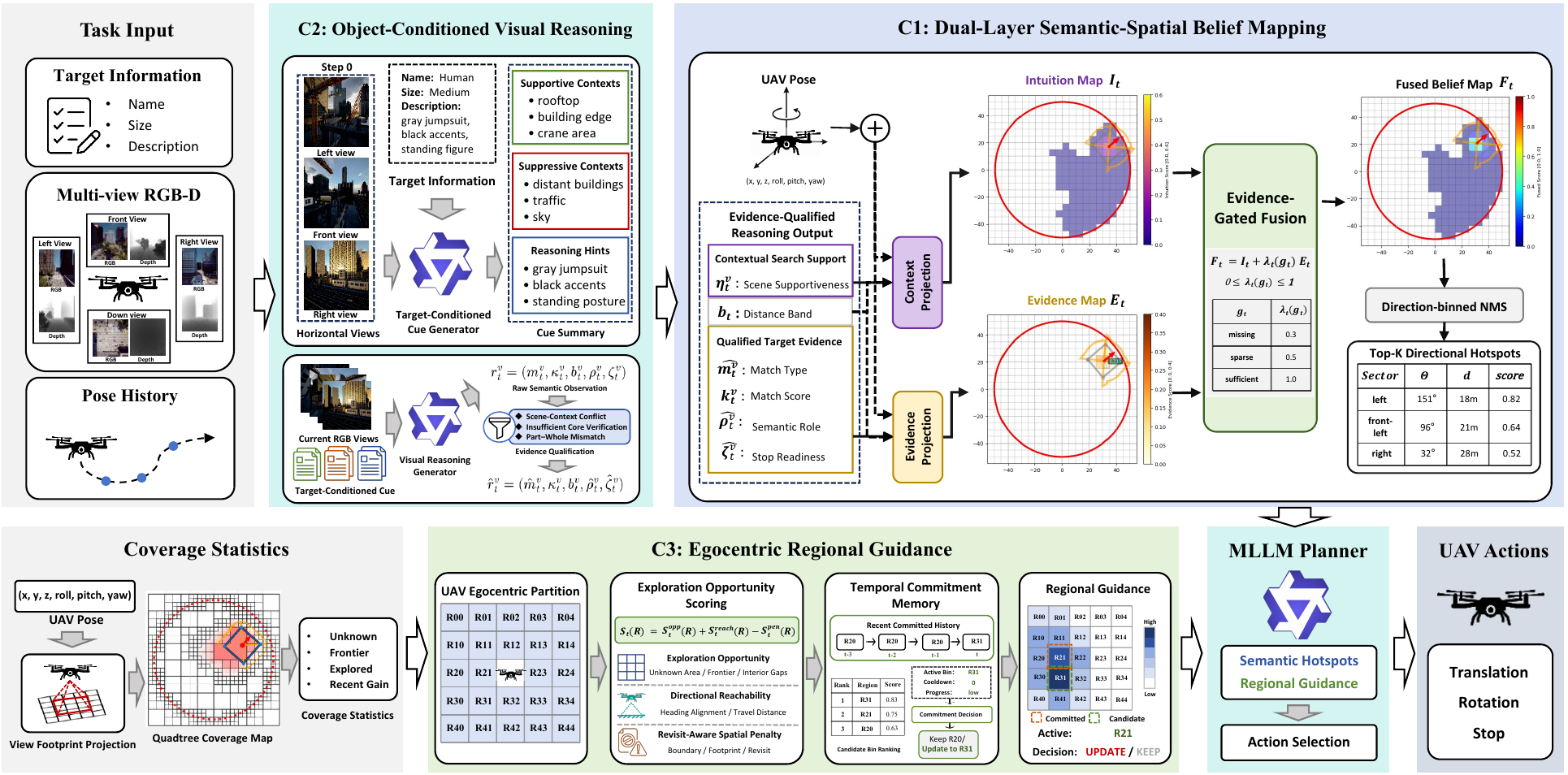}
\caption{Overview of the proposed framework. C2 performs target-conditioned visual reasoning with conservative evidence qualification. C1 projects and accumulates contextual support and qualified target evidence in separate intuition and evidence layers, which are fused into spatial belief hotspots. C3 converts quadtree coverage into egocentric directional guidance with temporal commitment. The planner jointly uses semantic belief and coverage-oriented guidance, while the shared UAV execution backbone filters infeasible actions and mitigates motion risks.}
\label{fig:system}
\end{figure*}

Fig.~\ref{fig:system} presents AeroBelief, which augments per-view VLM interpretation with persistent semantic-spatial belief and temporally stable exploration guidance. Object-conditioned visual reasoning constructs target-specific cues from the target specification and initial observations, while conservative evidence qualification restricts insufficiently supported target claims before spatial accumulation. Contextual search support and qualified target evidence are then projected into a shared aerial search grid and accumulated in separate intuition and evidence layers. Evidence-gated fusion combines the two layers into spatial belief hotspots for target-oriented planning.

In parallel, egocentric regional guidance projects quadtree coverage into UAV-centered, yaw-aligned bins and evaluates them according to exploration opportunity, reachability, and revisit-aware spatial efficiency. Regional scoring operates independently of semantic belief values, while temporal commitment stabilizes the selected direction across steps. The planner jointly uses fine-grained belief hotspots and coarse coverage-oriented guidance, while the shared execution backbone filters infeasible actions and mitigates motion risks.

\subsection{Object-Conditioned Visual Reasoning}
\label{sec:c2}

Frame-level VLM interpretations may contain target-specific evidence, weak contextual support, and visually similar distractors. We therefore introduce object-conditioned visual reasoning to interpret multi-view observations with respect to the episode target and conservatively qualify high-level target claims before spatial accumulation. The module first constructs a target-conditioned cue from the target specification and initial observations, and then produces structured semantic observations for the dual-layer belief map.

\subsubsection{Target-Conditioned Semantic Interpretation}

\begin{figure}[!t]
\centering
\includegraphics[width=\columnwidth]{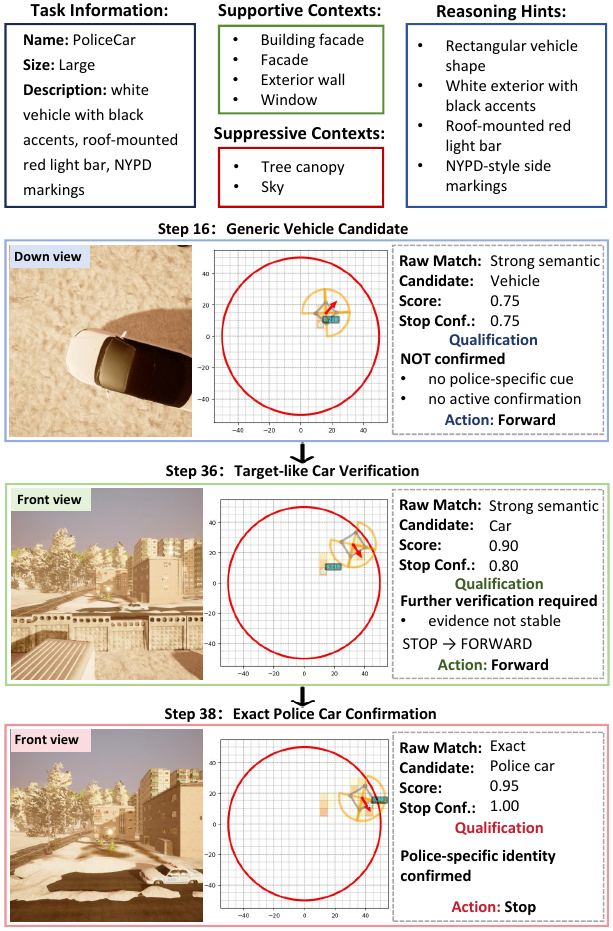}
\caption{Target-conditioned semantic interpretation and conservative evidence qualification in a representative episode. Target-specific cues distinguish contextual support from candidate target evidence, while conservative qualification restricts insufficiently supported claims before evidence-layer accumulation.}
\label{fig:c2_case}
\end{figure}

Given the target specification $g=(n_g,s_g,d_g)$ and the episode-initial front, left, and right RGB observations, we construct a target-conditioned cue $\mathbf{q}$ once per episode. The cue captures discriminative target attributes, compositional part--whole structure, range-dependent appearance, supportive and suppressive scene contexts, and the visual criteria required for target-level verification. It serves as an interpretation prior rather than an action policy and guides the VLM in distinguishing target-specific observations from weaker contextual compatibility.

At each semantic update step, the current multi-view observations are interpreted with respect to $\mathbf{q}$. For each processed view
$v\in\mathcal{V}$, the VLM produces a structured semantic judgment
\begin{equation}
r_t^v =
\left(
m_t^v,
\kappa_t^v,
b_t^v,
\rho_t^v,
\zeta_t^v
\right),
\label{eq:semantic_observation}
\end{equation}
where $m_t^v$ denotes the match type, $\kappa_t^v$ denotes the confidence score, $b_t^v$ denotes the estimated distance band, $\rho_t^v$ denotes the semantic role of the observation, and $\zeta_t^v$ denotes stop readiness. The match type takes one of four levels, namely \emph{exact}, \emph{strong semantic}, \emph{weak contextual}, and \emph{none}. Exact and strong semantic matches serve as candidate target-specific evidence, whereas weak contextual matches represent scene-level or component-level cues that may guide exploration but are insufficient for target confirmation. The semantic role further distinguishes the canonical target from target components, adjacent objects, and ambiguous matches.

Separately, we derive a scene-supportiveness score
$\eta_t^v\in[0,1]$ by relating the interpreted scene categories and visible objects to the supportive and suppressive context priors encoded in $\mathbf{q}$. Together, $\eta_t^v$ and the estimated distance band $b_t^v$ provide contextual search support for subsequent intuition-layer projection without constituting direct evidence of target presence.

\subsubsection{Conservative Evidence Qualification}

Despite target conditioning, the VLM may still overstate target presence when only contextual, partial, or visually similar cues are observed. Before spatial accumulation, we apply conservative evidence qualification to each structured observation and obtain
\begin{equation}
\hat r_t^v =
\left(
\hat m_t^v,
\kappa_t^v,
b_t^v,
\hat\rho_t^v,
\hat\zeta_t^v
\right),
\label{eq:qualified_observation}
\end{equation}
where the qualified match type $\hat m_t^v$, semantic role $\hat\rho_t^v$, and stop readiness $\hat\zeta_t^v$ may be revised when the available visual support is insufficient. The confidence score $\kappa_t^v$ and distance band $b_t^v$ are retained. The qualification procedure does not strengthen weak observations into target-specific evidence.

Qualification focuses on observations that assert \emph{exact} or \emph{strong semantic} matches associated with the canonical target. Three complementary conditions are examined. Scene-context conflict identifies claims dominated by suppressive contexts associated with potential false positives. Insufficient core verification identifies observations that do not satisfy the discriminative requirements encoded in the target-conditioned cue. Part--whole mismatch identifies responses triggered by an isolated target-like component or an adjacent object rather than the complete target.

When the available support is insufficient, the match is reclassified as \emph{weak contextual}, and its semantic role and stop-readiness signal are revised conservatively. Such an observation may still contribute to intuition-layer projection but is excluded from evidence-layer projection. This mechanism limits the persistent accumulation of insufficiently supported target claims while preserving useful contextual information for exploration. Fig.~\ref{fig:c2_case} illustrates this process in a representative episode.

\subsection{Dual-Layer Semantic-Spatial Belief Mapping}
\label{sec:c1}

The structured outputs produced by Sec.~\ref{sec:c2} remain frame-level signals and contain both contextual search support and qualified target-specific evidence. Accumulating these signals in a single map may conflate weak contextual cues with reliable evidence, while retaining only target-specific evidence would discard useful guidance before the target becomes visible. We therefore maintain two complementary layers over a shared world-frame aerial search grid. Each cell $c$ stores an intuition value $I_t(c)$ for context-guided exploration and an evidence value $E_t(c)$ for target approach and confirmation. The two layers receive different semantic inputs and follow different temporal update rules.

\subsubsection{Layer-Specific Belief Update}

For each view $v\in\mathcal{V}$, the qualified observation $\hat r_t^v$ is projected into the search grid using the current UAV pose, viewing direction, estimated distance band, and observable spatial support. For layer $L\in\{I,E\}$, the resulting view-specific contribution is expressed as
\begin{equation}
\Delta_{L,t}^v(c)
=
\operatorname{clip}
\left(
\alpha_L(\hat m_t^v)
\kappa_t^v
\Phi_L^v(c;b_t^v,p_t),
0,
\bar{\Delta}_L
\right),
\label{eq:view_projection}
\end{equation}
where $\alpha_L(\hat m_t^v)$ denotes the layer-specific semantic contribution associated with the qualified match type, $\bar{\Delta}_L$ bounds the contribution of a single observation, and $\Phi_L^v$ incorporates distance-band modulation and spatial attenuation according to the UAV pose, viewing geometry, and observable support. Horizontal observations are projected using depth-informed visibility and angular support, whereas the downward observation is associated with its ground footprint.

The positive contributions from all processed views are transformed into the same world coordinate frame and aggregated through a per-cell maximum,
\begin{equation}
\Delta_{L,t}(c)
=
\max_{v\in\mathcal{V}}
\Delta_{L,t}^v(c),
\qquad
L\in\{I,E\}.
\label{eq:multiview_aggregation}
\end{equation}
This max-belief aggregation avoids repeatedly adding overlapping observations to the same cell. Non-positive intuition signals, when present, are incorporated conservatively through smoothing rather than accumulated as persistent target evidence.

The intuition layer admits both target-specific and contextual semantic observations. Its admissible match set is
\begin{equation}
\mathcal{M}_I
=
\{
\text{exact},
\text{strong semantic},
\text{weak contextual}
\}.
\label{eq:intuition_match_set}
\end{equation}
Weak contextual observations can therefore guide exploration even when they are insufficient for target confirmation. When no explicit target match is available but the scene-supportiveness score $\eta_t^v$ is sufficiently high, a small bounded contextual prior is included in the intuition contribution and merged with other positive signals through the same max-belief update. This prior remains intentionally weak and cannot independently produce strong target belief.

The intuition update applies a local max-belief operation followed by global temporal forgetting,
\begin{equation}
I_t^{\mathrm{loc}}(c)
=
\min
\left\{
\bar I,
\max
\left(
I_{t-1}(c),
\Delta_{I,t}(c)
\right)
\right\},
\label{eq:intuition_local_update}
\end{equation}
\begin{equation}
I_t(c)
=
\begin{cases}
\gamma I_t^{\mathrm{loc}}(c),
&
\gamma I_t^{\mathrm{loc}}(c)
\ge
\tau_{\mathrm{clear}},
\\
0,
&
\text{otherwise},
\end{cases}
\label{eq:intuition_decay}
\end{equation}
where $\bar I$ denotes the upper bound of the intuition value, $\gamma$ is the temporal decay factor, and $\tau_{\mathrm{clear}}$ removes stale low-confidence cells. The local max operation retains the stronger of the previous belief and the current contribution, while global forgetting gradually removes outdated contextual support. The upper bound $\bar I$ limits the maximum intuition value stored in each cell.

The evidence layer is more selective. It admits only qualified target-specific observations whose match type belongs to
\begin{equation}
\mathcal{M}_E
=
\{
\text{exact},
\text{strong semantic}
\},
\label{eq:evidence_match_set}
\end{equation}
and whose confidence score exceeds the evidence-admission threshold $\tau_E$. Observations reclassified as \emph{weak contextual}, or failing to satisfy the confidence requirement, are excluded from evidence accumulation. The evidence update is defined as
\begin{equation}
E_t(c)
=
\min
\left\{
\bar E,
\max
\left(
E_{t-1}(c),
\Delta_{E,t}(c)
\right)
\right\},
\label{eq:evidence_update}
\end{equation}
where $\bar E$ denotes the maximum evidence value assigned to each cell. Unlike the intuition layer, the evidence layer does not decay within an episode. Qualified target evidence is therefore preserved when subsequent views temporarily lose target visibility because of motion, occlusion, or viewpoint change.

\subsubsection{Evidence-Gated Fusion and Hotspot Interface}

The intuition and evidence layers are integrated through evidence-gated fusion,
\begin{equation}
\begin{aligned}
F_t(c)
=
\operatorname{clip}
\Big(
&
\operatorname{clip}
\left(
I_t(c),0,\bar I_F
\right)
\\
&
+
\lambda_t(g_t)
\operatorname{clip}
\left(
E_t(c),0,\bar E_F
\right),
0,1
\Big),
\end{aligned}
\label{eq:evidence_gated_fusion}
\end{equation}
where $\bar I_F$ and $\bar E_F$ bound the respective contributions of the intuition and evidence layers during fusion. The evidence-gating state $g_t$ takes one of three values, namely \emph{missing}, \emph{sparse}, or \emph{sufficient}. The fusion weight $\lambda_t(g_t)$ increases with the availability and reliability of qualified target evidence. When evidence is missing or sparse, the fused map remains primarily influenced by contextual intuition. Once sufficient evidence accumulates, its contribution is increased to support target approach and confirmation.

This formulation preserves the distinct roles of contextual plausibility and target-specific evidence. Contextual cues guide exploration through the intuition layer but do not directly enter the evidence term. In the ablation variant without the evidence layer, the evidence term is removed and the fused representation reduces to an intuition-only map.

The fused belief map is further converted into a compact spatial interface for the planner. We extract the top-$K$ directional belief hotspots
\begin{equation}
\mathcal{H}_t
=
\{h_1,\ldots,h_K\}
\label{eq:belief_hotspots}
\end{equation}
from $F_t$ using direction-binned non-maximum suppression together with an explored-region penalty. Each hotspot summarizes a target-related spatial hypothesis through its directional sector, relative bearing, distance, and fused belief score. The hotspot set complements the rendered belief map by providing concise target-relevant spatial candidates. This fine-grained belief stream is jointly used with the egocentric regional guidance introduced in Sec.~\ref{sec:c3}.

\subsection{Egocentric Regional Guidance}
\label{sec:c3}

\begin{figure}[!t]
\centering
\includegraphics[width=\columnwidth]{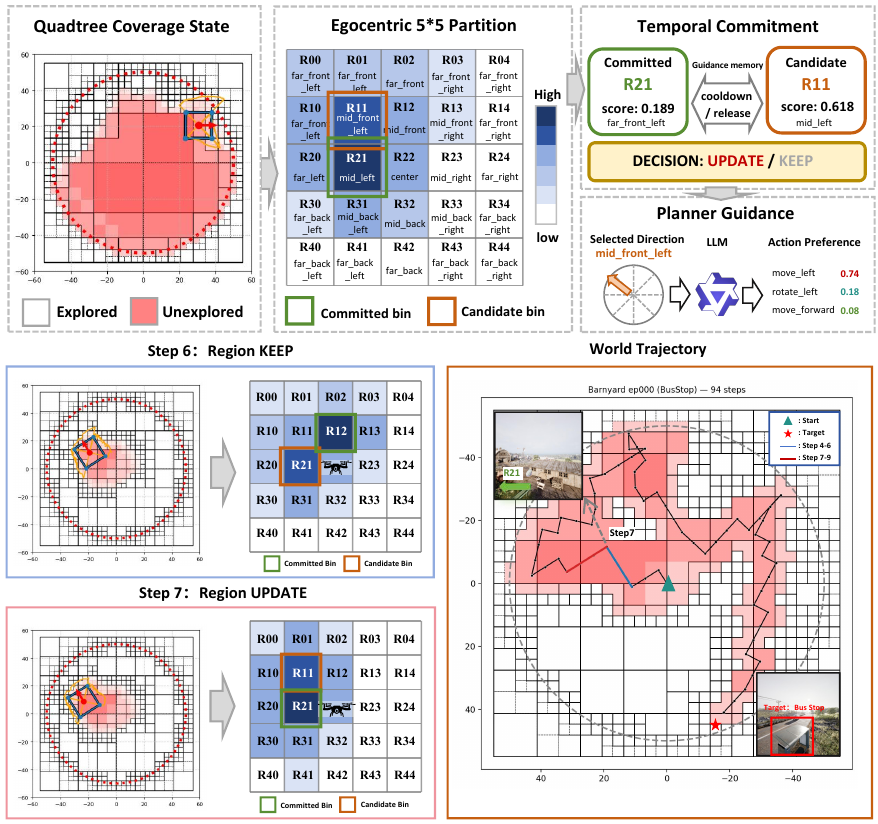}
\caption{Egocentric regional guidance with temporal commitment in a representative episode. Quadtree coverage is projected into UAV-centered, yaw-aligned directional bins to generate an instantaneous candidate. Temporal commitment retains or updates the selected direction across steps, and the committed direction is provided to the planner as coverage-oriented guidance. The bins represent egocentric directions rather than fixed world-space partitions.}
\label{fig:c3_case}
\end{figure}

The dual-layer belief map provides fine-grained target-related guidance, but long-horizon aerial navigation also benefits from systematic coverage when reliable semantic evidence remains sparse. We therefore introduce an egocentric regional guidance stream whose regional scoring operates independently of semantic belief values. Quadtree coverage is projected into UAV-centered, yaw-aligned directional bins and evaluated according to exploration opportunity, reachability, and revisit-aware spatial efficiency. Temporal commitment stabilizes the selected direction across steps. The resulting guidance complements the fine-grained semantic belief stream without directly prescribing the executed action.

\subsubsection{Egocentric Regional Proposal}

Let $p_t$ denote the current UAV pose and $\psi_t$ its yaw. Each quadtree coverage node is transformed into the current body frame and assigned to one of the $5\times5$ heading-relative bins
\begin{equation}
\mathcal{R}_t
=
\left\{
R_{ij}^{\,t}
\right\}_{i,j=0}^{4}.
\label{eq:egocentric_bins}
\end{equation}
The bins are reconstructed at each step according to the current $(p_t,\psi_t)$. Rows follow the forward--backward body axis, columns follow the left--right body axis, and the central bin contains the UAV. Each $R_{ij}^{\,t}$ therefore represents an egocentric directional region rather than a persistent world-space partition. As the UAV moves or rotates, the same bin index may correspond to different world locations.

Each bin is assigned a coverage-oriented score
\begin{equation}
S_t(R)
=
S_t^{\mathrm{opp}}(R)
+
S_t^{\mathrm{reach}}(R)
-
S_t^{\mathrm{pen}}(R),
\label{eq:regional_score}
\end{equation}
where $S_t^{\mathrm{opp}}$ measures remaining exploration opportunity, $S_t^{\mathrm{reach}}$ evaluates directional reachability, and $S_t^{\mathrm{pen}}$ accounts for spatial inefficiency. These terms consider unexplored-area support, frontier and interior gaps, heading alignment, travel distance, boundary exposure, valid spatial support, and visitation history. Semantic belief values are not used in the regional score.

The current coverage state yields an instantaneous candidate bin
\begin{equation}
R_t^{\mathrm{cand}}
=
\underset{R\in\mathcal{C}_t}{\arg\max}\;
S_t(R),
\label{eq:regional_candidate}
\end{equation}
where $\mathcal{C}_t\subseteq\mathcal{R}_t$ denotes the set of geometrically valid bins under the current search-boundary constraints. The candidate is recomputed at each step and represents the most promising instantaneous exploration direction.

\subsubsection{Temporal Commitment and Planner Coupling}

Directly using $R_t^{\mathrm{cand}}$ at every step may produce unstable exploration because small variations in the coverage state can repeatedly change the highest-scoring direction. We therefore maintain a cross-step temporal commitment that either retains the previous directional preference or updates it to the current candidate,
\begin{equation}
R_t^{\mathrm{com}}
=
\begin{cases}
R_{t-1}^{\mathrm{com}},
&
\mathrm{KEEP},
\\[2pt]
R_t^{\mathrm{cand}},
&
\mathrm{UPDATE}.
\end{cases}
\label{eq:region_commitment}
\end{equation}
The previous commitment is retained while it remains useful and the search continues to make progress. It is updated when the candidate provides a sufficiently stronger exploration opportunity or when low progress, obstacles, or repeated visitation indicate that the current direction has become stale. A minimum commitment duration and switching hysteresis further reduce short-term directional oscillation.

The commitment memory stores the selected egocentric bin identifier rather than a fixed world-space region. At each step, this identifier is interpreted with respect to the newly constructed body-frame grid, thereby preserving a relative directional preference as the UAV moves and rotates.

The committed bin is converted into a coarse body-frame directional cue for the planner. This cue represents an exploration preference rather than an exact metric waypoint. The planner jointly considers it with the fine-grained semantic belief hotspots and current visual observations, and may depart from the regional preference when stronger target evidence or motion constraints support another action. The shared execution backbone subsequently handles collision avoidance, boundary constraints, and other motion-safety requirements.

In the ablation variant without commitment memory, regional scoring and candidate generation remain active, but no regional state is retained across steps. The instantaneous candidate is therefore supplied directly to the planner at each step. Fig.~\ref{fig:c3_case} illustrates the regional proposal, temporal commitment, and planner coupling process.

\subsection{Planner-Level Integration}
\label{sec:planner_integration}

At each step, the planner receives qualified semantic observations, belief hotspots, committed regional guidance, recent pose--action history, and feasible-action information to produce a high-level discrete action proposal. A shared UAV execution layer then performs stop verification, collision avoidance, boundary and altitude handling, and fallback recovery for AeroBelief, the vanilla backbone, and all ablation variants.

\section{Experiments}
\label{sec:experiment}

\subsection{Experimental Setup}

We evaluate AeroBelief on the UAV-ON test set following the standard evaluation protocol~\cite{uavon}. Each episode permits up to 150 actions within a $50\,\mathrm{m}$ search radius, and success requires an explicit \texttt{stop} within $20\,\mathrm{m}$ of the target. Baseline results are taken from the corresponding publications because several recent methods lack public implementations. A locally deployed Qwen3-VL-8B is used for both visual semantic reasoning and MLLM-based high-level planning~\cite{qwen3vl}. Experiments run on four NVIDIA A800 GPUs with six parallel processes, each hosting two model instances. The full test-set evaluation takes approximately 13 days because every navigation step requires synchronous model inference.

Given this computational cost, ablation studies use a fixed 200-episode subset with 25 episodes from each of eight representative scenes. It includes four open scenes (Barnyard, WinterTown, CityStreet, and Neighborhood) and four cluttered scenes (NYC, Venice, UrbanJapan, and WesternTown). All variants use identical episodes and initial conditions, with invalid spawn configurations corrected using the same deterministic procedure.

\subsection{Compared Methods}

We compare AeroBelief with Random, CLIP-H, and AOA-F reported in UAV-ON~\cite{uavon}, together with AerialVLN~\cite{AerialVLN}, NaVid~\cite{navid}, OpenFly~\cite{openfly}, APEX~\cite{apex}, and OctMem-Agent~\cite{octmem}. Random selects actions without semantic guidance, CLIP-H uses image--text similarity as a heuristic search signal, and AOA-F combines vision-language scene interpretation with fixed-step aerial navigation. AerialVLN, NaVid, and OpenFly represent aerial vision-language navigation approaches, whereas APEX and OctMem-Agent are developed specifically for UAV-ON.

The Vanilla Backbone retains the shared visual interpretation, MLLM planner, quadtree coverage representation, and UAV execution layer, while disabling target-conditioned reasoning, dual-layer belief mapping, and egocentric regional guidance. It serves as a shared-backbone MLLM baseline under the same execution setting.

\subsection{Evaluation Metrics}
\label{sec:metrics}

We report Success Rate (SR), Oracle Success Rate (OSR), and Success weighted by Path Length (SPL). SR measures the percentage of episodes in which the UAV explicitly stops within $20\,\mathrm{m}$ of the target. OSR measures the percentage of all episodes in which the UAV enters the success radius before termination. Episodes terminated by collision are counted as failures. The OSR--SR gap reflects failures to convert target-vicinity discovery into successful stopping. SPL jointly measures navigation success and path efficiency.

\subsection{Main Comparison}

\begin{table*}[!t]
\caption{Overall and size-stratified performance on the UAV-ON benchmark. APEX is reported only in the Total column because its published results use seen/unseen splits rather than target-size groups.}
\label{tab:main}
\centering
\footnotesize
\renewcommand{\arraystretch}{1.08}
\setlength{\tabcolsep}{4.0pt}
\begin{tabular*}{\textwidth}{@{\extracolsep{\fill}}lcccccccccccc}
\hline
\multirow{2}{*}{Method}
& \multicolumn{3}{c}{Small}
& \multicolumn{3}{c}{Medium}
& \multicolumn{3}{c}{Large}
& \multicolumn{3}{c}{Total} \\
\cline{2-4}
\cline{5-7}
\cline{8-10}
\cline{11-13}
& SR$\uparrow$ & OSR$\uparrow$ & SPL$\uparrow$
& SR$\uparrow$ & OSR$\uparrow$ & SPL$\uparrow$
& SR$\uparrow$ & OSR$\uparrow$ & SPL$\uparrow$
& SR$\uparrow$ & OSR$\uparrow$ & SPL$\uparrow$ \\
\hline
Random
& 4.14 & 7.80 & 2.80
& 3.33 & 8.10 & 3.05
& 2.48 & 8.07 & 1.62
& 3.70 & 8.00 & 2.66 \\

CLIP-H
& 2.86 & 8.43 & 1.51
& 10.95 & 16.67 & \underline{7.17}
& 13.04 & 19.25 & 10.53
& 6.20 & 11.90 & 4.15 \\

AOA-F
& 4.45 & 16.38 & 1.61
& 10.48 & 17.62 & 6.36
& 14.29 & 21.74 & 10.66
& 7.30 & 17.50 & 4.06 \\

AerialVLN
& 4.13 & 19.87 & 4.01
& 4.76 & 17.62 & 4.90
& 1.86 & 19.25 & 1.86
& 3.90 & 19.30 & 3.72 \\

NaVid
& 10.02 & 29.55 & 4.41
& 11.43 & 26.44 & 5.95
& 17.39 & 30.73 & \textbf{15.03}
& 11.50 & 29.10 & 6.44 \\

OpenFly
& 12.40 & 26.23 & \underline{6.35}
& 10.00 & 20.48 & 4.19
& 13.04 & \underline{31.65} & 7.56
& 12.00 & 25.90 & 6.09 \\

APEX
& -- & -- & --
& -- & -- & --
& -- & -- & --
& 13.33 & 20.00 & \underline{10.14} \\

OctMem-Agent
& \underline{18.91} & \underline{29.57} & 6.03
& \underline{18.09} & \underline{26.66} & 6.27
& \textbf{23.60} & \textbf{31.67} & 7.80
& \underline{19.50} & \underline{29.30} & 6.37 \\

\hline

AeroBelief (Ours)
& \textbf{20.67} & \textbf{39.11} & \textbf{9.11}
& \textbf{25.31} & \textbf{32.72} & \textbf{12.03}
& \underline{20.16} & 26.05 & \underline{14.40}
& \textbf{21.61} & \textbf{35.57} & \textbf{10.62} \\
\hline
\end{tabular*}
\end{table*}

Table~\ref{tab:main} reports the overall and size-stratified results on the UAV-ON test set. AeroBelief achieves the best reported overall SR, OSR, and SPL among the compared methods, reaching 21.61\%, 35.57\%, and 10.62, respectively. Compared with the strongest previously reported result for each metric, AeroBelief improves SR by 2.11 percentage points, OSR by 6.27 percentage points, and SPL by 0.48. The gains are particularly evident for small and medium targets, where AeroBelief ranks first across all three metrics. For large targets, it achieves the second-highest SR and SPL, although its OSR remains lower than those of several prior methods.

The size-stratified results further reveal different navigation behaviors. Small targets exhibit the largest OSR--SR gap of 18.44 percentage points, suggesting that the UAV frequently reaches the target vicinity but does not always convert these opportunities into successful stopping. Medium targets achieve the highest SR among the three size groups, while large targets obtain the highest size-specific SPL despite their lower OSR. This suggests that successful large-target episodes remain relatively path-efficient once a reliable approach direction is established. Overall, the results support the effectiveness of combining persistent semantic-spatial belief with stable coverage-oriented guidance for aerial ObjectNav.

\begin{figure*}[!t]
\centering
\includegraphics[width=0.98\textwidth]{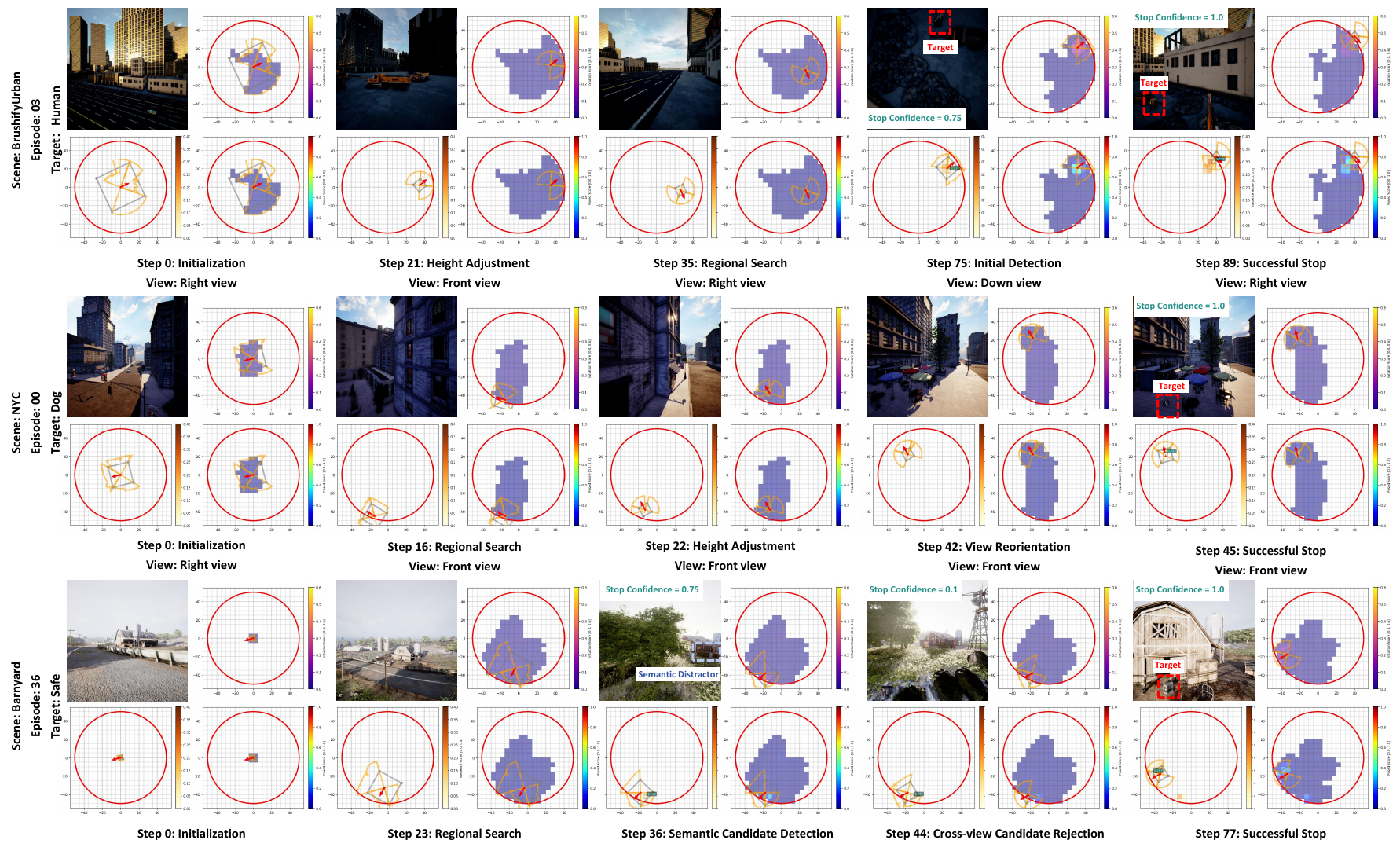}
\caption{Qualitative results from three successful episodes. Each example shows the UAV trajectory and the evolution of the intuition, evidence, and fused belief maps.}
\label{fig:qualitative}
\end{figure*}

Fig.~\ref{fig:qualitative} presents three representative successful episodes. During early exploration, the intuition layer maintains broad contextual belief over potentially relevant regions. As qualified target-specific observations accumulate, the evidence layer preserves more reliable spatial cues, and the fused belief gradually concentrates around the target. The examples further show that the UAV may continue regional search, adjust its viewpoint or altitude, and reject ambiguous candidates before sufficient evidence is obtained. The resulting trajectories illustrate how AeroBelief transitions from broad exploration to evidence-guided approach and explicit stopping.

\subsection{Ablation Study}

Table~\ref{tab:ablation} evaluates the proposed components on a fixed 200-episode subset. For each major module, we remove either the complete pathway or one of its key subcomponents. The belief variants remove both belief layers or only the evidence layer, the visual-reasoning variants remove the complete object-conditioned pathway or only conservative evidence qualification, and the regional-guidance variants remove the complete guidance module or only temporal commitment. Vanilla disables all proposed modules, while all variants share the same episodes, initial conditions, planner backbone, and UAV execution layer. We additionally report Collision and MaxCov, denoting the collision termination rate and the average maximum explored-area ratio across episodes, respectively.

\begin{table}[!t]
\caption{Ablation results on the fixed 200-episode subset.}
\label{tab:ablation}
\centering
\footnotesize
\renewcommand{\arraystretch}{1.08}
\resizebox{\columnwidth}{!}{
\begin{tabular}{@{}lccccc@{}}
\toprule
\multirow{2}{*}{Variant}
& \multicolumn{3}{c}{Navigation Performance}
& \multicolumn{2}{c}{Diagnostic Statistics} \\
\cmidrule(lr){2-4}
\cmidrule(l){5-6}
& SR$\uparrow$
& OSR$\uparrow$
& SPL$\uparrow$
& Coll.$\downarrow$
& MaxCov$\uparrow$ \\
\midrule
Vanilla & 11.00 & 22.50 & 5.95 & 39.50 & 23.5 \\
w/o Dual-Layer Belief Mapping
& 15.50 & 30.00 & 6.76 & 36.00 & 27.8 \\
w/o Evidence Layer
& 19.50 & \underline{35.00} & 9.10
& \underline{30.00} & \textbf{30.2} \\
w/o Object-Conditioned Reasoning
& 16.00 & 27.00 & 8.72 & 35.00 & 25.7 \\
w/o Evidence Qualification
& \underline{20.50} & \underline{35.00}
& \underline{9.80} & 30.50 & \underline{29.1} \\
w/o Regional Guidance
& 17.00 & 29.50 & 7.49 & 41.00 & 26.8 \\
w/o Commitment Memory
& 18.00 & 32.50 & 9.42 & 33.00 & 27.5 \\
AeroBelief
& \textbf{25.00} & \textbf{42.50}
& \textbf{13.36} & \textbf{28.00} & 29.0 \\
\bottomrule
\end{tabular}
}
\end{table}

AeroBelief achieves the highest SR, OSR, and SPL and the lowest collision rate, supporting the effectiveness of combining the three proposed modules. In all three module groups, removing the complete pathway causes a larger performance degradation than removing only its paired subcomponent. This pattern supports the contributions of semantic-spatial belief mapping, object-conditioned reasoning, and regional guidance, while the evidence layer, conservative evidence qualification, and temporal commitment provide additional gains. Compared with Vanilla, AeroBelief increases SR from 11.00\% to 25.00\% and reduces collision termination from 39.50\% to 28.00\%. Although removing the evidence layer produces a higher MaxCov, its lower SR and SPL indicate that broader coverage alone does not ensure effective target confirmation. Persistent target-specific evidence is therefore important for converting exploration into successful approach and stopping.

\section{Conclusion}

We presented AeroBelief, a dual-layer semantic-spatial belief mapping framework for vision-language aerial ObjectNav. It combines object-conditioned visual reasoning with conservative evidence qualification, evidence-gated belief fusion, and egocentric regional guidance with temporal commitment. These components provide complementary target-related belief and coverage-oriented guidance for planner-level decision making. Experiments on UAV-ON show that AeroBelief achieves the best reported overall SR, OSR, and SPL among the compared methods, while the ablation results support the contribution of each major component.

The current evaluation is limited to AirSim simulation, and sim-to-real transfer remains unvalidated. The gap between OSR and SR also indicates that reliable target confirmation and stopping remain important challenges. Future work will investigate real-world deployment, additional VLM backbones, improved stopping reliability, and stronger planner-level coordination between semantic belief and coverage-oriented guidance while preserving independent regional scoring.



\clearpage
\appendices

\begin{figure*}[!t]
\centering

{\LARGE\bfseries Supplementary Material\par}

\vspace{0.8em}

\includegraphics[width=0.96\textwidth]{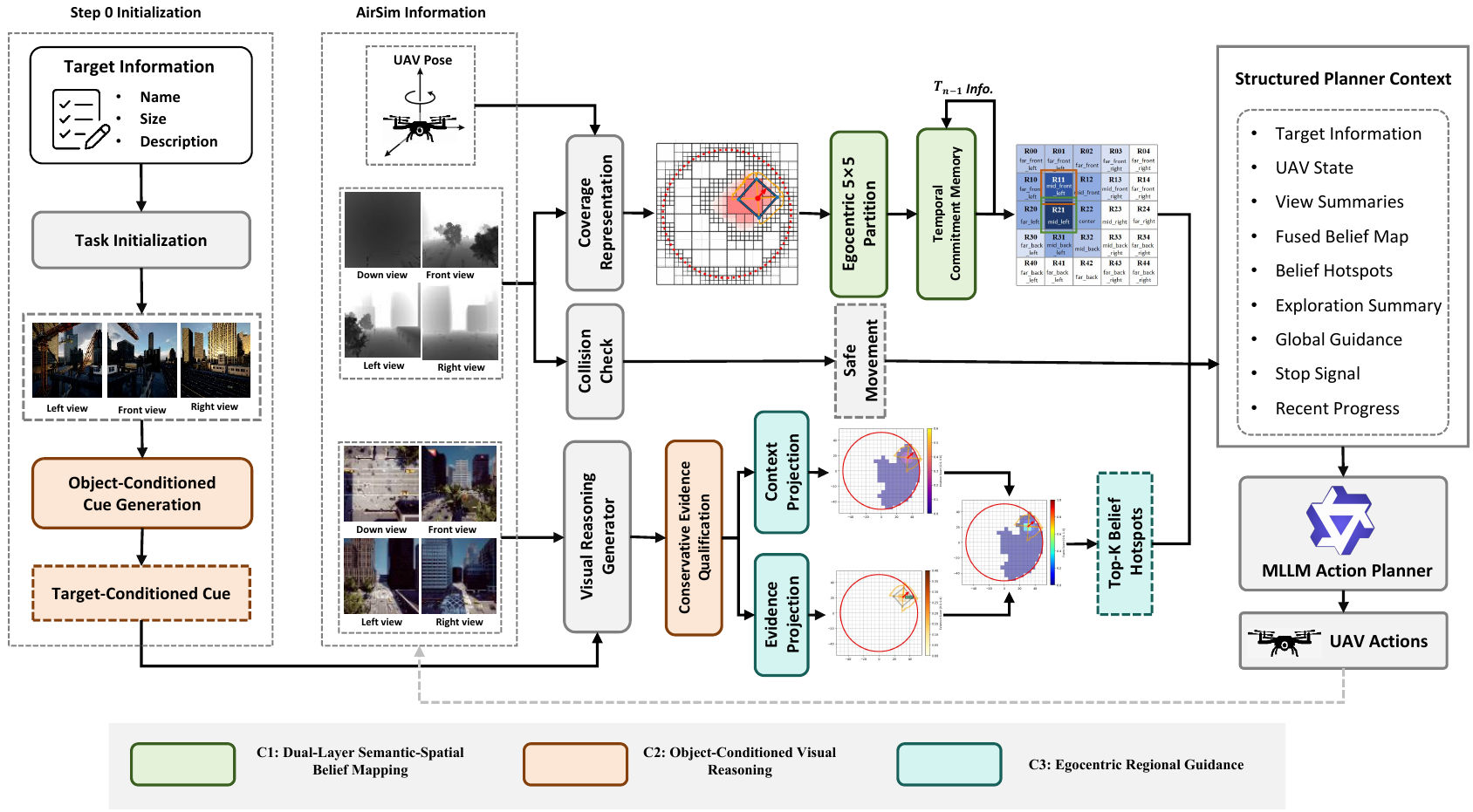}

\caption{Overview of the shared UAV backbone and proposed reasoning modules. All variants share the observation interface, generic per-view interpretation, quadtree coverage representation, planner, and execution pipeline, while C1--C3 denote the proposed enhancements.}
\label{fig:appendix_backbone}
\end{figure*}

\section{Shared Autonomous UAV Backbone}
\label{app:backbone}

\subsection{Observation and Action Interface}

At each timestep, the UAV acquires forward, left, right, and downward RGB views together with their corresponding depth observations. It also receives the current pose, flight height, recent pose--action history, and a target specification consisting of a name, size category, and natural-language description. Raw camera observations are processed by a shared generic per-view vision-language interpretation stage. The planner therefore operates on structured view summaries, spatial maps, the target specification, the current UAV state, and a compact history summary rather than directly processing raw RGB-D observations.

A quadtree coverage representation is incrementally maintained from the UAV observations. It summarizes explored and unexplored spatial structure and provides the common coverage information used by all evaluated variants. The egocentric regional scoring and temporal commitment mechanisms constructed on top of this representation are described separately in Appendix~\ref{app:quadtree}.

All variants use the same discrete action space,
\begin{equation}
\begin{aligned}
\mathcal{A}=\{&
\mathrm{forward},\,
\mathrm{left},\,
\mathrm{right},\,
\mathrm{ascend},\\
&
\mathrm{descend},\,
\mathrm{rotL},\,
\mathrm{rotR},\,
\mathrm{stop}
\}.
\end{aligned}
\end{equation}
Horizontal translations use a fixed step of $5\,\mathrm{m}$, vertical motions use a step of $2\,\mathrm{m}$, and rotations change the heading by $15^{\circ}$. The \emph{stop} action terminates the episode only when the shared verification conditions are satisfied. The action space, motion increments, action budget, search radius, and success criterion remain identical across AeroBelief, the vanilla backbone, and all ablation variants.

\subsection{Shared Execution Pipeline}

The action proposed by the planner is processed by a shared execution layer before being applied to the simulator. This layer evaluates motion feasibility using the current depth observations and UAV state. Motions with potential collision risks or search-boundary violations are rejected, while minimum-altitude protection prevents unsafe downward movement. The execution layer also applies target-size-aware altitude regulation. When the UAV is above the preferred height range associated with the target size, a descent may be preferred if the downward observation indicates sufficient clearance. When the proposed action cannot be executed safely, a feasible alternative is selected from the remaining action set.

The execution layer further provides recovery behavior for repeated boundary interactions, low-progress motion, and short-term oscillation. These mechanisms may redirect the UAV toward the interior of the search area or maintain corrective motion for several steps until it leaves an unsafe or unproductive state. This execution-level recovery is independent of the temporal commitment used by the egocentric regional guidance module.

Stopping follows the same verification procedure for all evaluated variants. A proposed stop is rejected when the current observations do not satisfy the shared target-verification criteria, and the search continues until sufficient target-specific support is available. Collision avoidance, boundary handling, target-size-aware altitude regulation, minimum-altitude protection, fallback recovery, and stop verification are shared by all evaluated variants and are not treated as contributions of the proposed semantic-spatial reasoning framework.

\subsection{Vanilla Search Behavior}

The vanilla backbone retains generic per-view visual interpretation, the MLLM planner, quadtree coverage representation, pose--action history, and the shared execution pipeline. It disables target-conditioned cue construction and conservative evidence qualification, dual-layer belief mapping, and egocentric regional guidance with temporal commitment. It therefore serves as a coverage-aware VLM--LLM baseline rather than a random-search baseline. AeroBelief, the vanilla backbone, and all ablation variants use identical episodes, action parameters, planning frequency, safety constraints, and stopping conditions, enabling controlled comparison of the proposed high-level reasoning components.

\section{Quadtree-Based Coverage Modeling}
\label{app:quadtree}

\subsection{Incremental Quadtree Coverage}

To compactly represent explored and unexplored space, we maintain an incremental quadtree over the search domain. The quadtree belongs to the shared navigation backbone and is updated for every evaluated variant independently of the semantic belief map. It records geometric observation coverage rather than target likelihood.

Let $\mathcal{Q}_t$ denote the quadtree at timestep $t$, and let
$\mathcal{L}(\mathcal{Q}_t)$ denote its current set of leaf nodes. Each leaf
$\ell\in\mathcal{L}(\mathcal{Q}_t)$ represents a world-frame spatial support
$\Omega(\ell)$ and maintains a coverage state $c_t(\ell)$. At each timestep,
the multi-view observations are projected onto the ground plane to obtain a
downward observation footprint $\mathcal{F}_t^{d}$ and a combined horizontal
visibility footprint $\mathcal{F}_t^{h}$. Their overlap with a leaf is measured by
\begin{equation}
\rho_t^{k}(\ell)
=
\frac{
\operatorname{area}
\left(
\Omega(\ell)\cap\mathcal{F}_t^{k}
\right)
}{
\operatorname{area}
\left(
\Omega(\ell)
\right)
},
\qquad
k\in\{d,h\},
\label{eq:coverage_overlap}
\end{equation}
where $\rho_t^{d}(\ell)$ and $\rho_t^{h}(\ell)$ denote the fractions observed
by the downward and horizontal views, respectively.

The two observation sources provide different forms of geometric coverage.
The downward view directly observes the underlying ground area, whereas
horizontal views cover larger projected regions with greater geometric
uncertainty. Their contributions are therefore incorporated through a
coverage-update operator
\begin{equation}
c_t(\ell)
=
\mathcal{U}
\left(
c_{t-1}(\ell),
\rho_t^{d}(\ell),
\rho_t^{h}(\ell)
\right),
\label{eq:coverage_update}
\end{equation}
which treats horizontal visibility conservatively relative to direct downward
observation.

Spatial refinement is performed adaptively. When an observation intersects
only part of a leaf and finer spatial resolution is required, the leaf is
subdivided into four children and the observation update is recursively
evaluated on the resulting cells. Large homogeneous regions therefore remain
compact, while observation boundaries and partially covered areas are
represented at finer resolution. Internal-node statistics are then updated
from their children to obtain the new quadtree $\mathcal{Q}_t$.

Algorithm~\ref{alg:quadtree_update} summarizes this incremental process.

\begin{algorithm}[t]
\caption{Incremental Quadtree Coverage Update}
\label{alg:quadtree_update}
\begin{algorithmic}[1]

\REQUIRE Previous quadtree $\mathcal{Q}_{t-1}$, UAV pose $p_t$,
downward footprint $\mathcal{F}_t^{d}$, horizontal footprint
$\mathcal{F}_t^{h}$

\ENSURE Updated quadtree $\mathcal{Q}_t$

\STATE $\mathcal{Q}_t \leftarrow \mathcal{Q}_{t-1}$

\STATE Transform $\mathcal{F}_t^{d}$ and $\mathcal{F}_t^{h}$
into the world frame using $p_t$

\STATE $\mathcal{V}_t \leftarrow
\left\{
\ell\in\mathcal{L}(\mathcal{Q}_t)
\mid
\Omega(\ell)\cap
\left(
\mathcal{F}_t^{d}\cup\mathcal{F}_t^{h}
\right)
\neq\emptyset
\right\}$

\FORALL{$\ell\in\mathcal{V}_t$}

    \STATE Compute $\rho_t^{d}(\ell)$ and $\rho_t^{h}(\ell)$
    using Eq.~\eqref{eq:coverage_overlap}

    \IF{$\operatorname{Refine}
    \left(
    \ell,\rho_t^{d}(\ell),\rho_t^{h}(\ell)
    \right)$}

        \STATE $\operatorname{Split}(\ell)$
        \STATE Recursively update the resulting child nodes

    \ELSE

        \STATE
        $c_t(\ell)
        \leftarrow
        \mathcal{U}
        \left(
        c_{t-1}(\ell),
        \rho_t^{d}(\ell),
        \rho_t^{h}(\ell)
        \right)$

    \ENDIF

\ENDFOR

\STATE Update internal-node coverage statistics from their children

\RETURN $\mathcal{Q}_t$

\end{algorithmic}
\end{algorithm}

\subsection{Coverage-Guided Regional Reasoning}

At each step, the leaves of $\mathcal{Q}_t$ are transformed into the body frame defined by the current UAV pose $p_t$ and yaw $\psi_t$, and aggregated into a $5\times5$ set of UAV-centered, yaw-aligned bins,
\begin{equation}
\mathcal{R}_t
=
\left\{
R_{ij}^{\,t}
\right\}_{i,j=0}^{4}.
\end{equation}
Rows follow the forward--backward body axis, columns follow the left--right axis, and the central bin contains the UAV. Because the bins are reconstructed from the current pose at every step, they represent relative directions rather than persistent world-space regions.

Each bin receives a coverage-oriented score
\begin{equation}
S_t(R)
=
S_t^{\mathrm{opp}}(R)
+
S_t^{\mathrm{reach}}(R)
-
S_t^{\mathrm{pen}}(R),
\end{equation}
where the three terms summarize remaining exploration opportunity, directional reachability, and revisit-aware spatial penalties. They consider unknown-area ratio, frontier and interior gaps, heading alignment, travel distance, boundary exposure, valid spatial support, and visitation history. Semantic belief values are not used in this score.

The instantaneous candidate is selected as
\begin{equation}
R_t^{\mathrm{cand}}
=
\underset{R\in\mathcal{C}_t}{\arg\max}\;
S_t(R),
\end{equation}
where $\mathcal{C}_t$ contains the geometrically valid bins under the current search boundary. To reduce frequent switching between similarly scored directions, the module maintains a committed egocentric bin,
\begin{equation}
R_t^{\mathrm{com}}
=
\begin{cases}
R_{t-1}^{\mathrm{com}},
&
\mathrm{KEEP},
\\[2pt]
R_t^{\mathrm{cand}},
&
\mathrm{UPDATE}.
\end{cases}
\end{equation}
The commitment is retained while the selected direction remains useful and is updated when a stronger candidate emerges or the current direction becomes stale. A minimum commitment duration and switching hysteresis further reduce short-term oscillation.

The commitment stores a relative bin identifier rather than a fixed world-space region, and is interpreted in the current body-frame grid after the UAV moves or rotates. The resulting direction is supplied to the planner as coarse coverage-oriented guidance rather than an exact waypoint. Without temporal commitment, the instantaneous candidate is passed directly to the planner at each step.

\section{Prompt Design and Structured Reasoning Interface}

\subsection{Target-Conditioned Visual Reasoning}

\begin{figure}[!t]
\centering
\fbox{
\begin{minipage}{0.92\columnwidth}
\small

\textbf{Episode-Level Cue Generation}

\textbf{Input}

Target name, size category, natural-language description,
and initial front, left, and right RGB views.

\textbf{Instruction}

\begin{itemize}
    \item Identify discriminative target attributes.
    \item Describe part--whole structure and verification cues.
    \item Summarize appearance across viewing ranges.
    \item Identify supportive and suppressive scene contexts.
    \item List likely confusers and rejection cues.
\end{itemize}

\textbf{Output}

Structured target-conditioned cue $\mathbf{q}$.

\vspace{0.5em}
\hrule
\vspace{0.5em}

\textbf{Step-Level Per-View Visual Reasoning}

\textbf{Input}

Current RGB view $I_t^v$, target specification $g$,
and episode-level cue $\mathbf{q}$.

\textbf{Instruction}

\begin{itemize}
    \item Classify the target-match type and estimate confidence.
    \item Estimate the coarse distance band.
    \item Distinguish contextual support from target-specific evidence.
    \item Summarize visible scene and object content.
    \item Assess stop readiness.
\end{itemize}

\textbf{Output}

Structured semantic observation $r_t^v$.

\end{minipage}
}
\caption{Simplified prompt structure for target-conditioned visual reasoning.}
\label{fig:prompt_structure}
\end{figure}

The visual reasoning interface contains two complementary prompting stages. At the beginning of each episode, a target-conditioned cue is generated once from the target specification and the initial forward, left, and right RGB observations. The cue-generation prompt is not action-oriented. It summarizes the visual knowledge required for subsequent interpretation, including discriminative attributes, part--whole structure, range-dependent appearance, supportive and suppressive scene contexts, likely confusers, and target-level verification cues.

The resulting cue $\mathbf{q}$ is reused throughout the episode as episode-level interpretation context. At each semantic update step, the current camera views are processed independently using the per-view reasoning prompt together with the target specification and $\mathbf{q}$. The prompt estimates the target-match type and confidence, coarse distance band, contextual and target-specific support, visible scene content, and stop readiness. It produces the structured observation $r_t^v$ defined in Sec.~\ref{sec:c2} without regenerating the episode-level cue.

The resulting structured observations are subsequently processed by conservative evidence qualification before entering the semantic-spatial belief map. The cue supports visual interpretation only. It neither prescribes navigation actions nor enters the planner as a standalone instruction.

\subsection{Structured Planner Interface}

\begin{figure}[!t]
\centering
\fbox{
\begin{minipage}{0.92\columnwidth}
\small

\textbf{Structured Planner Input}

\vspace{0.3em}

\textbf{Map Inputs}\\
Rendered fused belief map; rendered quadtree coverage map.

\vspace{0.3em}

\textbf{Task Context}\\
Target specification; episode progress.

\vspace{0.3em}

\textbf{UAV Context}\\
Pose; heading; height; feasible actions; recent motion history.

\vspace{0.3em}

\textbf{Visual Context}\\
Qualified per-view summaries; target-match signals; stop-readiness signals.

\vspace{0.3em}

\textbf{Semantic Belief Context}\\
Belief summary; top-$K$ belief hotspots.

\vspace{0.3em}

\textbf{Coverage Context}\\
Exploration statistics; coverage progress; recent search gain.

\vspace{0.3em}

\textbf{Regional Guidance}\\
Committed egocentric direction and associated action preference.

\vspace{0.5em}
\hrule
\vspace{0.5em}

\textbf{Output}\\
High-level discrete action proposal
$\rightarrow$
shared execution backbone
$\rightarrow$
executed UAV action.

\end{minipage}
}
\caption{Simplified structured interface of the navigation planner. The planner receives rendered spatial maps and organized textual context rather than raw RGB-D observations.}
\label{fig:planner_interface}
\end{figure}

The navigation planner operates on a structured multimodal interface that combines rendered spatial maps with organized textual context rather than raw RGB-D observations. The map inputs include the fused semantic belief map and quadtree coverage map.

The textual context is organized into several functional groups. Task context specifies the target and current episode progress. UAV context describes the current pose, heading, height, feasible actions, and recent motion history. Visual context summarizes qualified per-view interpretations, target-match information, and stop-readiness signals. Semantic belief context exposes the current belief summary and top-$K$ hotspots, while coverage context reports exploration statistics and recent search progress. Regional guidance provides the committed egocentric direction and associated action preference produced by Sec.~\ref{sec:c3}. The target-conditioned cue is not inserted verbatim into the planner prompt, but influences planning through the structured visual observations and semantic-spatial reasoning outputs.

Given this structured context, the planner produces a high-level discrete action proposal from the shared action space. The proposal is subsequently processed by the common execution backbone for stop verification, collision avoidance, boundary and altitude handling, and fallback recovery. Because this execution backbone is shared across AeroBelief and all ablation variants, it provides a common execution setting for evaluating the proposed high-level reasoning components.

\section{Supplementary Analysis}

\subsection{Failure Mode Analysis}

\begin{table}[!t]
\caption{Episode termination modes and stop precision on the test set.}
\label{tab:failure_analysis}
\centering
\footnotesize
\setlength{\tabcolsep}{6pt}
\begin{tabular}{lc}
\toprule
Termination / Metric & Percentage \\
\midrule
Collision & $21.48\%$ \\
Step limit & $27.50\%$ \\
Explicit stop & $51.03\%$ \\
\midrule
Stop precision & $42.36\%$ \\
\bottomrule
\end{tabular}
\end{table}

We further analyze episode termination behavior to identify the main remaining failure modes. Explicit \texttt{stop} accounts for $51.03\%$ of episode terminations, while $27.50\%$ reach the maximum action budget and $21.48\%$ terminate because of collision. These results show that the remaining failures arise from both incomplete exploration and inaccurate termination decisions. Stop precision denotes the fraction of explicit-stop episodes that satisfy the success criterion.

Episodes reaching the step limit generally correspond to cases in which the target is not sufficiently discovered or verified within the available search budget. Collision termination instead reflects the difficulty of sustaining reliable exploration in geometrically complex environments despite the shared execution safeguards.

Explicit stopping represents the largest termination category, but its precision is only $42.36\%$. This indicates that semantic confidence is not always well aligned with the geometric success criterion. The UAV may recognize the target with high visual confidence and issue \texttt{stop} while remaining outside the required success radius. A substantial fraction of the remaining failures therefore arises not from complete target-discovery failure, but from insufficient calibration between semantic evidence, estimated proximity, and successful termination.

This issue is particularly relevant to physically large targets, whose apparent visual scale may provide a misleading indication of distance. The following subsection examines this size-dependent stopping behavior.

\subsection{Target-Size Sensitivity}

\begin{figure}[!t]
\centering
\includegraphics[width=\columnwidth]{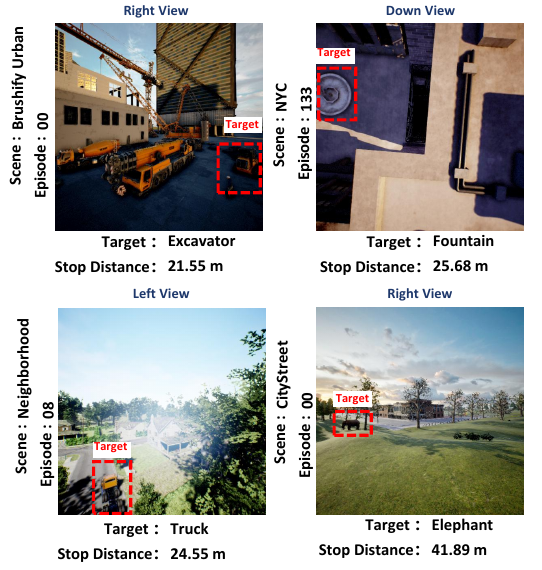}
\caption{Representative premature stops involving large targets. Strong visual saliency may trigger termination outside the $20\,\mathrm{m}$ success radius.}
\label{fig:size_failure}
\end{figure}

The size-specific results show that large targets do not consistently outperform medium-sized targets despite their greater visual saliency. Inspection of failed episodes suggests that this behavior is associated more strongly with distance-sensitive stopping than with initial target discovery. Large objects can often be recognized from relatively long distances, while their physical scale makes visual proximity estimation more ambiguous.

Stopping in the current framework depends on structured visual reasoning signals, including target-match strength, coarse distance estimation, and stop readiness. These signals are inferred primarily from visual appearance rather than direct target-distance measurements. For a physically large object, a substantial image footprint does not necessarily indicate that the UAV is already within the success radius. However, strong semantic visibility and large apparent scale may lead the VLM to produce an optimistic distance estimate or high stop readiness before the UAV has approached sufficiently closely. The system may therefore recognize the target but terminate outside the required success radius.

Fig.~\ref{fig:size_failure} presents representative premature-stop cases involving large targets. In each example, the target is visually salient in the current observation, but the UAV stops beyond the $20\,\mathrm{m}$ success threshold. The cases span different object categories and viewpoints, suggesting that the behavior is not confined to a single scene or target appearance. They illustrate a recurring ambiguity between \emph{visual saliency} and \emph{metric proximity}, since large targets may become visually dominant before the UAV is geometrically close enough for successful termination.

These observations suggest that many large-target failures are stopping-calibration failures rather than pure search failures. The target may be discovered early and recognized with high semantic confidence while the remaining distance is still underestimated. Future work could incorporate more reliable geometric distance estimation or target-size-aware stopping criteria to better separate recognition confidence from metric proximity.

\end{document}